# ASSAMESE-ENGLISH BILINGUAL MACHINE TRANSLATION

Kalyanee Kanchan Baruah[1], Pranjal Das[1], Abdul Hannan[1] and Shikhar Kr Sarma[1]

[1]Department of Information Technology, Gauhati University, Guwahati, Assam

*ABSTRACT*

*Machine translation is the process of translating text from one language to another. In this paper, Statistical Machine Translation is done on Assamese and English language by taking their respective parallel corpus. A statistical phrase based translation toolkit Moses is used here. To develop the language model and to align the words we used two another tools IRSTLM, GIZA respectively. BLEU score is used to check our translation system performance, how good it is. A difference in BLEU scores is obtained while translating sentences from Assamese to English and vice-versa. Since Indian languages are morphologically very rich hence translation is relatively harder from English to Assamese resulting in a low BLEU score. A statistical transliteration system is also introduced with our translation system to deal basically with proper nouns, OOV (out of vocabulary) words which are not present in our corpus.*

## 1. INTRODUCTION

Machine Translation (MT), subfield of computational linguistics plays a major role in men machine communication as well as men-men communication. It is used to translate text from one natural language to another. Translation is not only used to translate literary works from one language to another it is also used for translation of scientific documents, technical manuals, documents and memoranda etc. In this paper, we are showing translation from two language Assamese-English and vice-versa. As most of the official works done in English there is a strong requirement of MT system to make it convenient for common people. Here Statistical Machine Translation is used which is a part of Machine Translation that strives to use machine learning paradigm towards translating text. Statistical Machine Translation consists of Language Model (LM), Translation Model(TM) and Decoder. In this paper bilingual Assamese-English Machine Translation Model has been developed by using Moses decoder. Other translation tools like IRSTLM are used to develop language model and GIZA++ to align the words of source text to target one. The measuring metric BLEU (Bilingual Evaluation Understudy) is used for automatic evaluation that is quick and language independent [4]. We also tried to compare the BLEU scores obtain while using English-Assamese and Assamese-English parallel corpora and found a significant difference in BLEU scores. The difference in BLEU score tells us how the morphology of Indian language like Assamese varies from English. We are trying to deal with OOV (out of vocabulary) words which were either ignored or dropped by statistical translation system by introducing Transliteration. Transliteration is the process of preserving the original sound by mapping the word from a source to target language using the pronunciation instead of translation of meaning. Statistical MT techniques have not so far been widely explored for Indian languages. We would like to explore as much as we can and see to what extent these models can contribute to the huge ongoing MT efforts in the country. Assamese (অসমীয়া) language is one of the foremost languages spoken in the north eastern part of India, mainly in the state of Assam, by over 19 million native speakers. It is one of the official languages of Assam. It is also spoken in various parts of other north-eastern states of India. The Assamese script is very much similar to Bengali script. Assamese is written from left to right, just like English [9].There are various works going on in the field of Natural Language Processing in Assamese. Assamese is a computationally underdeveloped language and research on NLP is still at its developing stage. Some of the researches are going on POS tagging in Assamese language using CRF and TRF [10], Automatic identification of MWE (Multi words Expressions) [11].

## 2. RELATED STUDY

There are many projects going on machine translation in India. Some of them are shown below
*Anglabharati* : Anglabharati is a pattern directed rule based system developed in IIT Kharagpur [12]. It produces translation between English to Hindi and was developed for a Public Heath Campaigns.

*Anusaaraka* : Anusaaraka is the MT system which is divided into two modules. The first part is based on language knowledge while the other part is domain specific, based on word and statistical knowledge [13].

*MaTra* : MaTra is an indicative English to Hindi Machine Translation System which is fully automatic. The approach taken by MaTra is transfer-based [14].

*UGSC MAT* : It translates English Text to Kannada.

Some projects are going on Natural Language Processing in Assam on Machine Translation, WordNet etc.

## 3. TOOLS FOR IMPLEMENTING THE SYSTEM

### 3.1. Moses

Moses is a statistical machine translation system designed and developed by Philipp Koehn and Hieu Hoang at the University of Edinburgh. It is a software that is used in translation of sentences from one language to another. Moses requires a parallel corpus or translated texts (for e.g. English and Assamese) that are used in training the system. Moses works in tandem with SRILM or IRSTLM to develop the Language Model and Giza++ to develop the Translation Model. Nowadays, with the increasing demand for high quality translations, Moses is steadily becoming the most popular translation system [1].

### 3.2. Giza++

Giza++ is a toolkit designed by Franz Josef Och. It is an extension of the program GIZA, developed by the Statistical Machine Translation team. Giza++ is used to develop the Translation Model of our system. Giza++ implements different models like HMM and it also helps in word alignment. It also contains source for *mkcls* tool which is used to train word classes by maximum likelihood criterion [2].

### 3.3. IRSTLM

IRSTLM is a language modeling tool that is used to develop the language model for machine translation system. It estimates, represent and computes statistical language models. It is used to develop the language model for the target language. It is licensed under open source LGPL [3]

### 3.4. BLEU

The BLEU (Bi-Lingual Evaluation Understudy) toolkit helps us to determine the quality of our translation. We can test how good our translation is by translating the text and then running the BLEU script on it. A simple BLEU scoring script is *multi-bleu.perl* [1][4].

## 4. METHODOLOGY

Statistical Machine Translation system requires parallel corpus of source and target language pair for translation of sentences. We have used about 2500 parallel English and Assamese sentences related to tourism domain. There are three main component of the SMT system –

1) Language Model (LM)
2) Translation Model (TM)
3) Decoder

Language Model computes the probability of target language sentences. Translation Model calculates the probability of target sentences given the source sentence and the Decoder maximizes the probability of the translated text [5].The architecture of our system is shown in the following figure:

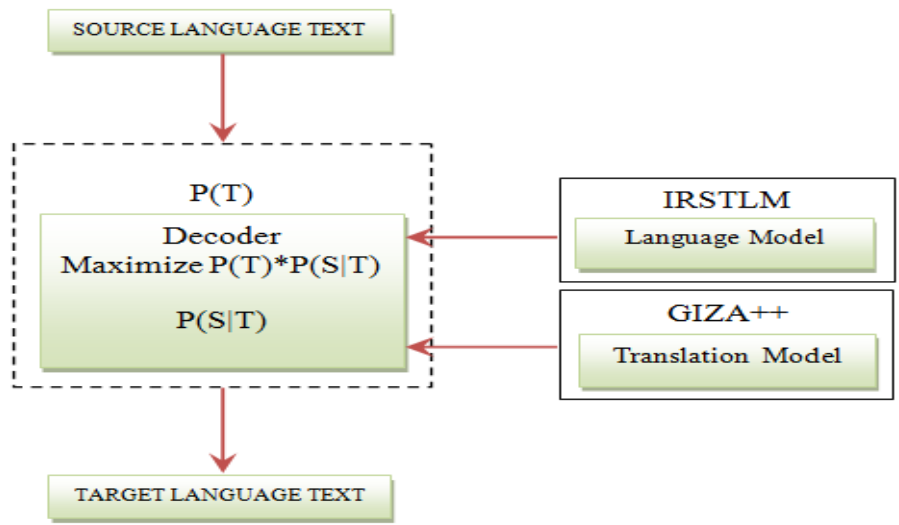

Fig. 1: Architecture of the SMT system

### 4.1. Language Model

The purpose of the language model is to develop fluent output by calculating the probability of a sentence. The probability is computed using *n-gram* model. Language Model can be considered as computation of the probability of single word given all of the words that precede it in a sentence [7]. The Language Model is developed using IRSTLM tool. The Language Model breaks the probability of a sentence P(S) as the probability of individual words P(w) shown in Eq.1

$$P(s) = P(w_1, w_2, w_3,....., w_n)$$

$$=P(w_1)P(w_2/w_1)P(w_3/w_1w_2)P(w_4/w_1w_2w_3)...P(w_n/w_1w_2...w_{n-1}) \quad (1)$$

For calculating the probability of a sentence, we are first required to calculate the probability of a word, given the sequence of word preceding it[16]. An *n-gram* model simplifies the task by approximating the probability of a word given all the previous words. An *n-gram* of size 1 is known as a *unigram*; size 2 is a *bigram*, also known as *digram*; size 3 is a *trigram*; size 4 is a *four-gram* and size 5 or more is simply called *n-gram* [5].

Suppose for a large amount of corpus we have the bigram probabilities given in the following table

Table 1. Bigram probabilities.

| Eat on | .16 | Eat Thai | .03 | <start> I | .25 | Want some | .04 |
|---|---|---|---|---|---|---|---|
| Eat some | .06 | Eat breakfast | .03 | <start> I'd | .06 | Want Thai | .01 |
| Eat lunch | .06 | Eat in | .02 | <start> Tell | .04 | To eat | .26 |
| Eat dinner | .05 | Eat Chinese | .02 | <start> I'm | .02 | To have | .14 |
| Eat at | .04 | Eat Mexican | .02 | I want | .32 | To spend | .09 |
| Eat a | .04 | Eat tomorrow | .01 | I would | .29 | To be | .02 |
| Eat Indian | .04 | Eat dessert | .007 | I don't | .08 | British food | .60 |
| Eat today | .03 | Eat British | .007 | I have | .04 | British restaurant | .15 |
| Eat French | 0.2 | Eat American | .002 | Want to | .65 | British cuisine | .01 |
| Eat Tomorrow | 0.5 | Eat Japanese | .003 | Want a | .05 | Chinese food | .01 |

Then, the probability of a sentence "*I want to eat Chinese food*" is P(I want to eat British food)
  = P(I|<start>) P(want | I) P(to | want) P(eat | to) P(Chinese | eat) P(food | Chinese)
  = .25*.32*.65*.26*.02*.01 = .000013

## 4.2. Translation Model

The Translation Model calculates similarity between input and output.The Translation Model computes the probability of source sentence '*S*', for a given target sentence '*T*' i.e. P(S|T). It is trained from the probability of target-source language pair[16]. Computation of translation model probabilities at sentence level is quite impossible, so the process is broken down into smaller units for e.g. words or phrases and their probabilities are learned[8].The target translation of source sentence is thought of as being generated from source word by word or phrases. For example, using the notation (T/S) to represent an input sentence S and its translation T, a sentence is translated as given below

(উদয়পুৰ এখন বিখ্যাত চহৰ| Udaipur is a famous city)

Udaipur is a famous city

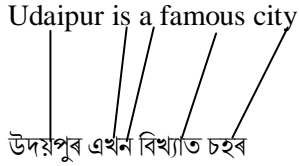

উদয়পুৰ এখন বিখ্যাত চহৰ

### 4.3. Decoder

Decoding is the process of finding a target translation sentence (English/Assamese) for a source sentence (Assamese/English) using translation model and language model. Decoding is a search problem which maximizes the translation and language model probability. The words and phrases are chosen which have maximum probability of being the translated translation [6]. Search for sentence T is performed that maximizes P(S|T)i.e.

$$\text{Pr}(S,T) = \mathit{argmax}\, P(T)\, P(S|T) \qquad (2)$$

### 4.4. Preparing Data For Training

Before training the data, we have to perform some tasks. We have to prepare the data for training the translation system. The following steps are performed:

- Tokenizing: We have to tokenize the data, so that spaces are inserted between words and punctuation [1].
- Truecasing: We then set the case of the first word in each sentence and this is done by truecasing the data [1].
- Cleaning: Then we have to clean the data i.e. remove empty lines, extra spaces and some lines that are too short or too long [1].

We can now train our translation model. After training is completed, a *moses.ini* is generated which is used in running the Moses decoder. We can improve our translation system by tuning the data. Tuning the system again results in a *moses.ini* file. We can now use this file to run the decoder[1].

### 4.5. Corpus Preparation

A corpus is a collection of written text in linguistics research, used to do statistical analysis and hypothesis testing [15]. Our corpus consists of parallel text in English and Assamese languages. The corpus is mostly based on travel and tourism in India. A sample of the corpus is shown in the following table.

Table 2. Assamese-English Parallel Corpus.

| English Sentence | Assamese sentence |
|---|---|
| Lallgarh Palace was the most completely integrated example of Indo-Saracenic architecture. | ভাৰত-আৰাকান ভাস্কৰ্যৰ আত সুন্দৰ সমাহাৰৰ উৎকৃষ্ট নিদৰ্শন হ'ল এই লালগড় প্ৰসাদা |
| Camel Breeding Farm is Just 8 km away from the city, at the govt. run camel breeding farm, you discover a lot about the Ship of the Desert . | উট প্ৰজনন ফাৰ্মখন মহানগৰৰ মাত্ৰ৮ কিলো'মিটাৰ দূৰত্বত অৱস্থিত৷ চৰকাৰীভাৱে পৰিচালনা কৰা এই ফাৰ্মত আপুনি আৱিষ্কাৰ কৰিব পাৰিব মৰুভূমিৰ জাহাজৰ নানান নজনা কথা |
| Jama Masjid is the largest mosque in India | জামা মছবজদ ভাৰতৰ ভিতৰতআটাইতকৈ ডাঙৰ মছজিদ . |
| The present temple known as the Maha Meru Prasad was rebuilt about 25 years ago on the location of the original shrine in accordance with ancient plans. | বৰ্তমানৰ মহা মুৰু প্ৰসাদ নামৰ মন্দিৰটো২৫ বছৰ পূৰ্বে পুৰণি মন্দিৰৰ স্থানত প্ৰাচীন আৰ্হি মতে সজোৱা হৈছিল৷ |

## 4.6. Transliteration

As we are provided with a very small amount of corpus, it is not possible to cover all the words. Some proper nouns, i.e. names of people, places etc. are not translated as they are not in our corpus. So we tried to introduce the transliteration system into our SMT system. Transliteration is defined as the process of automatically mapping a character from one language with the character from another language such that it preserves the pronunciation of the original source word. For example,

(k-u-m-a-r) ->ক-ু-ম-া-ৰ

We have used a Perl script to retrieve those words which are not translated or which are not in our corpus. In the Perl file, we have stored all the Assamese characters along with their corresponding English transliterations and executed the Perl script with Moses. A sample test result is shown in the following table

Table 3. Using transliteration in our system.

| Source Sentence | Before Transliteration | After Transliteration |
|---|---|---|
| কানাদা এখন বিশাল দেশ | কানাদা is a vast country | kanada is a vast country |
| Panaji is a city | Panaji হৈছে এটা চহৰ | পানাজি হৈছে এটা চহৰ |

In the first example "কানাদা এখন বিশাল দেশ", the actual spelling of the word "কানাদা" is "কানাডা". As we have taken the letter 'd' as the transliteration letter for 'দ', instead of 'ড' so we cannot get the correct word. Likewise we cannot obtain the correct word for some sentences both in English and Assamese. This is because we are concentrating only on the phonetic transcription. We are thinking to tryout some other methods which will solve this problem.

## 5. RESULTS AND EVALUATION OF BLEU SCORE

We have trained the system and as a result a *moses.ini* configuration file is generated. We have used this file to run Moses. Again, we have tuned the system to get a better output. As our system is built with a very limited parallel corpus (about 2500 sentences), some of our translations are quite rough but understandable. We have trained a system for Assamese to English translation and another system for English to Assamese translation. The test results are explained in the following tables:

Table 4. Assamese to English Translation.

| Input Assamese Sentence | Output English sentence |
| --- | --- |
| জয়পুৰ ৰাজস্থানৰ এখন বিখ্যাত চহৰ। | Jaipur is a famous city of Rajasthan . |
| তাজমহল আগ্ৰাত অৱস্থিত। | The Taj Mahal , is located in the heart of the Agra . |
| হায়দৰাবাদৰ পৰা ইণ্ডিয়ান এয়াৰলাইন্স, জেট এয়াৰওৱেজ আৰু চাহাৰা এয়াৰলাইন্সৰ বিমান সেৱাৰ ব্যৱস্থা আছে। | Hyderabad from Indian Airlines , Jet airways and Sahara Airlines with frequent flights air services . |
| কেওলাদেও ঘন ৰাষ্ট্ৰীয় উদ্যান বিশ্বৰ অতিকৈ ধুনীয়া জলপক্ষী অভয়াৰণ্যৰ ভিতৰত এখন। | the world famous for its proximity to the Keoladeo Ghana National Park is one of the finest water-bird sanctuaries in |
| গুৱাহাটী অসমৰ ৰাজধানী। | Guwahati is connected by the capital of the State . |
| মমতাজ মহল শ্বাহজাহানৰ পত্নী আছিল। | Mumtaz Mahal Shahjahan , |
| অন্ধ্ৰপ্ৰদেশ ভাৰতৰ এখন অন্যতম ৰাজ্যৰ ভিতৰত এক। | Andhra Pradesh is one of the state of one of India . |
| জামা মছজিদ শ্বাহজাহানে নিৰ্মাণ কৰিছিল। | Jama Masjid built by Shahjahan . |

International Journal on Natural Language Computing (IJNLC) Vol. 3, No.3, June 2014Table 5. English to Assamese Translation.

| Input English Sentence | Output Assamese sentence |
|---|---|
| Delhi is the capital of India | ভাৰতৰ ৰাজধানী দিল্লী |
| Jodhpur city is the second largest city of the state | যোধপুৰ চহৰখন দ্বিতীয় সৰ্ববৃহৎ নগৰী |
| Jama Masjid is the largest mosque in India | জামা মছজিদ ভাৰতৰ ভিতৰত আটাইতকৈ ডাঙৰ মছজিদ। |
| City Palace is a synthesis of Mughal and Rajasthani architecture | চিটি পেলেচ মোগল আৰু ৰাজস্থানী স্থাপত্য বিদ্যাৰ সংমিশ্ৰণ। |
| The British declared full sovereignty over the Falkland Islands the following year . | ডি সম্পূৰ্ণ সাৰ্বভৌমত্বৰ ঘোষণা কৰিছিল। ইয়াৰ পিছৰ বছৰত বৃটিছসকলে ফকলেণ্ড দ্বীপসমূহৰ ওপৰত |
| Assam is a beautiful place | অসম এখন ধুনীয়া ৰাজ্য |
| Bharatpur museum is a major source to have a date with the past royal glory of the place | ভৰতপুৰ সংগ্ৰাহালয় এক প্ৰধান উৎস , য'ৰ অতীতৰ ৰাজকীয় গৌৰৱৰ বিষয়ে জানিব পাৰি। |

We have evaluated the BLEU score for our translated sentences using the BLEU (Bilingual Evaluation Understudy) toolkit. BLEU score determines the efficiency of our system i.e. how good our translation system is. The following table shows the scores of Assamese to English and English to Assamese translation.

Table 6. Bleu Score.

| Source/Target | Bleu Score |
|---|---|
| Assamese/English | 9.72 |
| English/Assamese | 5.02 |

Our BLEU scores are quite low because of the lesser availability of parallel corpus. Also, though we have used the same corpus for training Assamese-English system and English-Assamese system, we are getting a vast difference in BLEU scores. This is because Indian languages are morphologically very rich and translation from English to Assamese is relatively harder than Assamese to English translation.

## 6. CONCLUSIONS

We have tried to build both Assamese to English and English to Assamese translation using the same parallel corpus. We are successful in our job to quite an extent but still we feel that there is

80

International Journal on Natural Language Computing (IJNLC) Vol. 3, No.3, June 2014much improvement to be made. About 2500 sentences are used for training both the system. But this is quite less because for a better translation system, the size of parallel corpus should be large. As our system is phrase-based, increasing the number of sentences will increase the quality of the translated text. We will work on this problem and try to increase the number of sentences into our system. Also, we will try to put our system into web-based portal rather than displaying the output in the terminal. We will also try to build a more developed translation system so that we can enhance our translation result and contribute a lot to the field of Assamese Natural Language Processing.

## ACKNOWLEDGEMENTS

The authors are thankful to the *Department of Information Technology, Gauhati University* for providing us the corpus, which helped us in building the SMT system and also the *Department of Natural Language Processing, Gauhati University* for their immense support.## REFERENCES

[1]  Statistical Machine Translation System User Manual and Code Guide", Available: http://www.statmt.org//moses/manual/manual.pdf.
[2]  F.J.Och., "GIZA++: Training of statistical translation models", Available: http://fjoch.com/GIZA++.html.
[3]  "IRSTLM", Available: http://hlt.fbk.eu/en/irstlm.
[4]  Kishore Papineni, Salim Roukos, Todd Ward and Wei-Jing Zhu, "Bleu: a method for Automatic Evaluation of Machine Translation", "Proceedings of the 40[th] Annual Meeting of the Association for Computer Linguistics (ACL)" Philadelphia, July 2002 pp. 311-318.
[5]  N.Sharma, P.Bhatia, V.Singh, "English to Hindi Statistical Machine Translation", June 2011.
[6]  P.F. Brown, S. De.Pietra, V. D. Pietra and R. Mercer, "The mathematics of statistical machine translation: parameter estimation". "Journal Computational Linguistics", vol. 10, no.2, June 1993.
[7]  "Machine Translation", Available: http://faculty.ksu.edu.sa/homiedan/Publications/Machine%20Translation.pdf.
[8]  D. D. Rao, "Machine Translation A Gentle Introduction", RESONANCE, July 1998.
[9]  "Assamese Language", Available: http://en.wikipedia.org/wiki/Assamese_Language.
[10] R.M.K Sinha, K Shivaraman, A Agarwal, R Jain, A Jain, "ANGLABHARATI: a multilingual machine aided translation project on translation from English to Indian languages.
[11] Akshar Bharati, Vineet Chaitanya, Amba P. Kulkarni, Rajeev Sangal, G Umamaheshwara Rao " Anusaaraka Machine Translation in stages.".
[12] Jayprasad J Hegde, Chandra Shekhar, Ritesh Shah, Sawani Bade, "MaTra A Practical Approach to Fully-Automatic Indicative English.
[13] "Text corpus", Available:http://en.wikipedia.org/Text_corpus.
[14] Aneena George,"English To Malayalam Statistical Machine Translation System".81

**Authors**

**Kalyanee Kanchan Baruah:** M.Tech Student,
Department of Information Technology,
Gauhati University

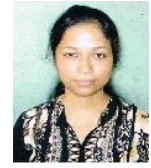

**Pranjal Das:** M.Tech Student,
Department of Information Technology,
Gauhati University

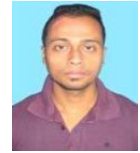

**Abdul Hannan:** PhD Scholar,
Department of Information Technology,
Gauhati University

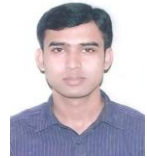

**Shikhar Kr. Sarma:** Head of the Department,
Department of Information Technology,
Gauhati University

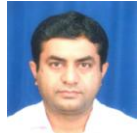